\documentclass[10pt,twocolumn,letterpaper]{article}

\usepackage{wacv}
\usepackage{times}
\usepackage{epsfig}
\usepackage{graphicx}
\usepackage{amsmath}
\usepackage{amssymb}
\usepackage{booktabs}

\usepackage{multirow}
\usepackage{makecell}
\usepackage{xcolor}

\usepackage[accsupp]{axessibility}  

\def\wacvPaperID{736} 

\wacvalgorithmstrack   

\wacvfinalcopy 

\ifwacvfinal
\usepackage[breaklinks=true,bookmarks=false]{hyperref}
\else
\usepackage[pagebackref=true,breaklinks=true,colorlinks,bookmarks=false]{hyperref}
\fi

\usepackage[capitalize]{cleveref}
\crefname{section}{Sec.}{Secs.}
\Crefname{section}{Section}{Sections}
\Crefname{table}{Table}{Tables}
\crefname{table}{Tab.}{Tabs.}

\begin{document}

\newcommand{\methodname}[1]{\textsc{Perceiver-VL}}

\newcommand{\vl}[1]{vision-and-language}
\newcommand{\Vl}[1]{Vision-and-language}
\newcommand{\pcv}[1]{\textsc{Perceiver}}
\newcommand{\pcvio}[1]{\textsc{Perceiver-IO}}

\newcommand{\frozen}[1]{Frozen-in-Time}
\newcommand{\hero}[1]{HERO}
\newcommand{\mmt}[1]{MMT}
\newcommand{\alvnet}[1]{AVLNET}
\newcommand{\clipbert}[1]{ClipBERT}

\newcommand{\visualbert}[1]{VisualBERT}
\newcommand{\lxmert}[1]{LXMERT}
\newcommand{\uniter}[1]{UNITER-Base}
\newcommand{\oscar}[1]{OSCAR-Base}
\newcommand{\vinvl}[1]{VinVL-Base}
\newcommand{\pixelbert}[1]{Pixel-BERT-R50}
\newcommand{\vilt}[1]{ViLT-B/32}
\newcommand{\vit}[1]{ViT-B/32}
\newcommand{\allinone}[1]{All-in-one-B}

\newcommand{\cc}[1]{CC}
\newcommand{\imagenet}[1]{ImageNet-21k}
\newcommand{\imagenetone}[1]{ImageNet-1k}
\newcommand{\webvid}[1]{Webvid}
\newcommand{\htm}[1]{HT100M}
\newcommand{\coco}[1]{COCO}
\newcommand{\tvqa}[1]{TVQA}
\newcommand{\vg}[1]{VG}
\newcommand{\sbu}[1]{SBU}
\newcommand{\flickr}[1]{Flickr}
\newcommand{\gqa}[1]{GQA}
\newcommand{\vgqas}[1]{VG-QAs}
\newcommand{\vgqa}[1]{VGQA}
\newcommand{\oi}[1]{OI}

\newcommand{\msrvtt}[1]{MSRVTT}
\newcommand{\didemo}[1]{DiDeMo}
\newcommand{\lsmdc}[1]{LSMDC}
\newcommand{\activitynet}[1]{ActivityNet}
\newcommand{\tgifqa}[1]{TGIF-QA}
\newcommand{\msrvttqa}[1]{MSRVTT-QA}

\newcommand{\vqa}[1]{VQAv2}
\newcommand{\nlvr}[1]{NLVR$^2$}
\newcommand{\flickrthirty}[1]{Flickr30k}

\definecolor{bittersweet}{rgb}{1.0, 0.44, 0.37}


\title{\methodname{}: Efficient Vision-and-Language Modeling \\ with Iterative Latent Attention}

\newcommand*\samethanks[1][\value{footnote}]{\footnotemark[#1]}

\author{
Zineng Tang\thanks{equal contribution} \quad Jaemin Cho\samethanks \quad Jie Lei \quad Mohit Bansal \\
UNC Chapel Hill\\
{\tt\small\{terran, jmincho, jielei, mbansal\}@cs.unc.edu}
}

\maketitle
\thispagestyle{empty}

\begin{abstract}

We present \methodname{}, a \vl{} framework that efficiently handles high-dimensional multi-modal inputs such as long videos and text.
Powered by the iterative latent cross-attention of Perceiver, our framework scales with linear complexity, in contrast to the quadratic complexity of self-attention used in many state-of-the-art transformer-based models. 
To further improve the efficiency of our framework, we also study applying LayerDrop on cross-attention layers and introduce a mixed-stream architecture for cross-modal retrieval.
We evaluate \methodname{} on diverse video-text and image-text benchmarks, where \methodname{} achieves the lowest GFLOPs and latency while maintaining competitive performance.
In addition, we also provide comprehensive analyses of various aspects of our framework, including pretraining data, scalability of latent size and input size, dropping cross-attention layers at inference to reduce latency, modality aggregation strategy, positional encoding,  and weight initialization strategy.\footnote{Our code and checkpoints are available at: \url{https://github.com/zinengtang/Perceiver_VL}
}

\end{abstract}

\section{Introduction}
\label{sec:intro}

During the past several years, there has been an increasing interest in \vl{} (VL) learning.
Many recent models~\cite{tan2019lxmert,lu2019vilbert,chen2019uniter,sun2019videobert,li2020hero,tang2021decembert,lei2021less} adopt the transformer~\cite{vaswani2017attention} architecture to encode \vl{} inputs.
These methods have improved the performance of various tasks, such as text-based image/video retrieval~\cite{chen2015microsoft,plummer2015flickr30k,xu2016msr,rohrbach2015dataset,caba2015activitynet,anne2017localizing,lei2020tvr,lei2021qvhighlights} and visual question answering~\cite{antol2015vqa,jang2017tgif,xu2017video,lei2018tvqa,yu2019activitynet}.
However, the transformer is based on the self-attention module~\cite{vaswani2017attention} with a quadratic computational cost in relation to its input length. This makes it difficult for models to process high-dimensional data, such as long videos.

To this end, we propose \methodname{}, an end-to-end \vl{} architecture that efficiently handles high-dimensional multi-modal inputs. 
\methodname{} is built on the iterative latent cross-attention of the recently proposed \pcv{}~\cite{jaegle2021perceiver1,jaegle2021perceiver2}. 
Concretely, we map a multi-modal input array of size $M$ to a latent array of size $N$ with cross-attention.
This changes the computational complexity of the attention modules from $O(M^2)$ to $O(NM)$.
Since \vl{} models often handle very long input arrays (\eg, $M >1000$),\footnote{The Frozen-in-time~\cite{bain2021frozen} visual encoder takes video inputs of frame length 8, frame size 224$\times$224, and patch size 16$\times$16. The resulting input length $M {=} (224/16)^2 \times 8 {=} 1568$, which is much larger than a typical latent array size $N{=}128$ in \methodname{}.} this greatly improves the efficiency for \vl{} tasks.
To further enhance the efficiency of our framework,
we also study reducing the number of cross-attention layers based on LayerDrop~\cite{fan2019reducing} and using a mixed-stream architecture for cross-modal retrieval tasks. 
By varying the number of cross-attention layers that take the most computation, we allow users to flexibly control the latency at inference.
The mixed-stream architecture combines the widely used single-stream and multi-stream architectures and improves the retrieval performance of the multi-stream architecture with a minimum increase in latency.

We evaluate \methodname{} on various video-text (\msrvtt{}, \didemo{}, \lsmdc{}, \activitynet{}, \tgifqa{}, \msrvttqa{}) and image-text (\flickrthirty{}, \vqa{}, \nlvr{}) tasks. 
Overall, \methodname{} achieves performance competitive to recent \vl{} models,
while maintaining significantly higher efficiency with the lowest GFLOPs and latency.
We demonstrate that \methodname{} scales more efficiently than the transformer-based architecture with respect to video length and frame size.
In addition, we show that our method also allows for flexible adaptions to further improve its efficiency:
(i) Decreasing the size of a latent array during finetuning reduces the computation significantly, with only a minimal accuracy drop.
(ii) Mixed-stream architecture achieves a reasonable accuracy-latency trade-off, with higher accuracy than multi-stream and lower latency than single-stream for text-to-video retrieval.
(iii) We apply LayerDrop during training. This allows users to control the latency by reducing the number of cross-attention layers at inference, again with only minimal accuracy drop. Training with LayerDrop also improves model performance.
Lastly, we conduct comprehensive ablation studies, including weight initialization, pretraining dataset, and modality aggregation methods.
We find it helpful to initialize the parameters from ViT~\cite{dosovitskiy2020image} and CLIP~\cite{radford2021learning} and pretrain jointly on video-text and image-text pairs.
We do not find a significant difference in combining two input modalities in a joint or separate attention module, and whether to use
learned~\cite{gehring2017convolutional,vaswani2017attention} or Fourier~\cite{stanley2007compositional,mildenhall2020nerf,jaegle2021perceiver1} positional encoding for the latent array.

Our contributions can be summarized as:
(1) We propose \methodname{}, an efficient \vl{} framework with linear scalability, on-demand depth reduction, and mixed-stream retrieval architecture.
(2) We demonstrate that our framework achieves significantly higher efficiency than recent transformer-based models on various \vl{} benchmarks, with overall competitive performance.
(3) We provide a comprehensive analysis of the efficiency, architectural components, and training strategies of our framework.
We hope that our research allows the community to use the highly efficient framework for diverse \vl{} tasks and inspires future research.

\section{Related Work}

\subsection{Efficient Transformers}

Many research works have proposed to reduce the quadratic computation complexity of self-attention in transformers~\cite{vaswani2017attention} based on different methods, including
hashing~\cite{kitaev2020reformer},
sparse attention~\cite{child2019generating,beltagy2020longformer,tay2020sparse},
kernel trick~\cite{katharopoulos2020transformers}, low-rank key/value projection~\cite{wang2020linformer}, blockwise attention~\cite{qiu2019blockwise},
past memory compression \cite{rae2019compressive},
and inducing point methods~\cite{lee2019set}.
Unlike these methods, \pcv{}~\cite{jaegle2021perceiver1} proposes using
iterative cross-attention to map an input array to a smaller latent array and apply self-attention to the latent array, which makes the computation scale linearly.
\pcvio{}~\cite{jaegle2021perceiver2} adds a decoder to \pcv{} to allow the model to tackle various downstream tasks with structured prediction.

To our knowledge, the iterative cross-attention of \pcv{} for multi-modal inputs has only been studied on audio-video autoencoding task~\cite{jaegle2021perceiver2}.
In this work, we present \methodname{}, which extends the \pcv{} framework in \vl{} domain.
We also evaluate \methodname{} on diverse video-text and image-text benchmarks and conduct extensive experiments to analyze its efficiency.
In addition, we introduce new techniques including cross-attention drop and mixed-stream architecture for cross-modal retrieval. 

\subsection{Vision-and-Language Pretraining}
Large-scale pretraining of transformers~\cite{vaswani2017attention,devlin2018bert} has achieved great success in natural language processing
\cite{liu2019roberta,yang2019xlnet,lan2019albert,dong2019unified,song2019mass,raffel2020exploring,clark2020electra}.
Following this success,
image-text~\cite{tan2019lxmert,lu2019vilbert,chen2019uniter,li2020unimo,zhou2020unified,li2020unicoder,lei2021less,cho2021unifying,radford2021learning,jia2021scaling} and
video-text~\cite{sun2019videobert,zhu2020actbert,li2020hero,tang2021decembert,zellers2021merlot,zellers2022merlot,tan2020vokenization,tang2021vidlankd}
multi-modal transformers have achieved improvements in various \vl{} tasks~\cite{antol2015vqa,chen2015microsoft,xu2016msr,zhou2017towards,lei2018tvqa}.
Such models take visual and textual inputs and are pretrained on large image-text/video-text pairs, with multi-modal masked language modeling and vision-text matching objectives with a transformer architecture~\cite{vaswani2017attention}.
One prominent issue with such models is that they are hard to scale because of the quadratic computation cost of self-attention layers.
In this work, we propose a new \vl{} pretraining framework that scales more efficiently than the transformer-based frameworks.

\begin{figure*}[t]
  \centering
\includegraphics[width=0.85\textwidth]{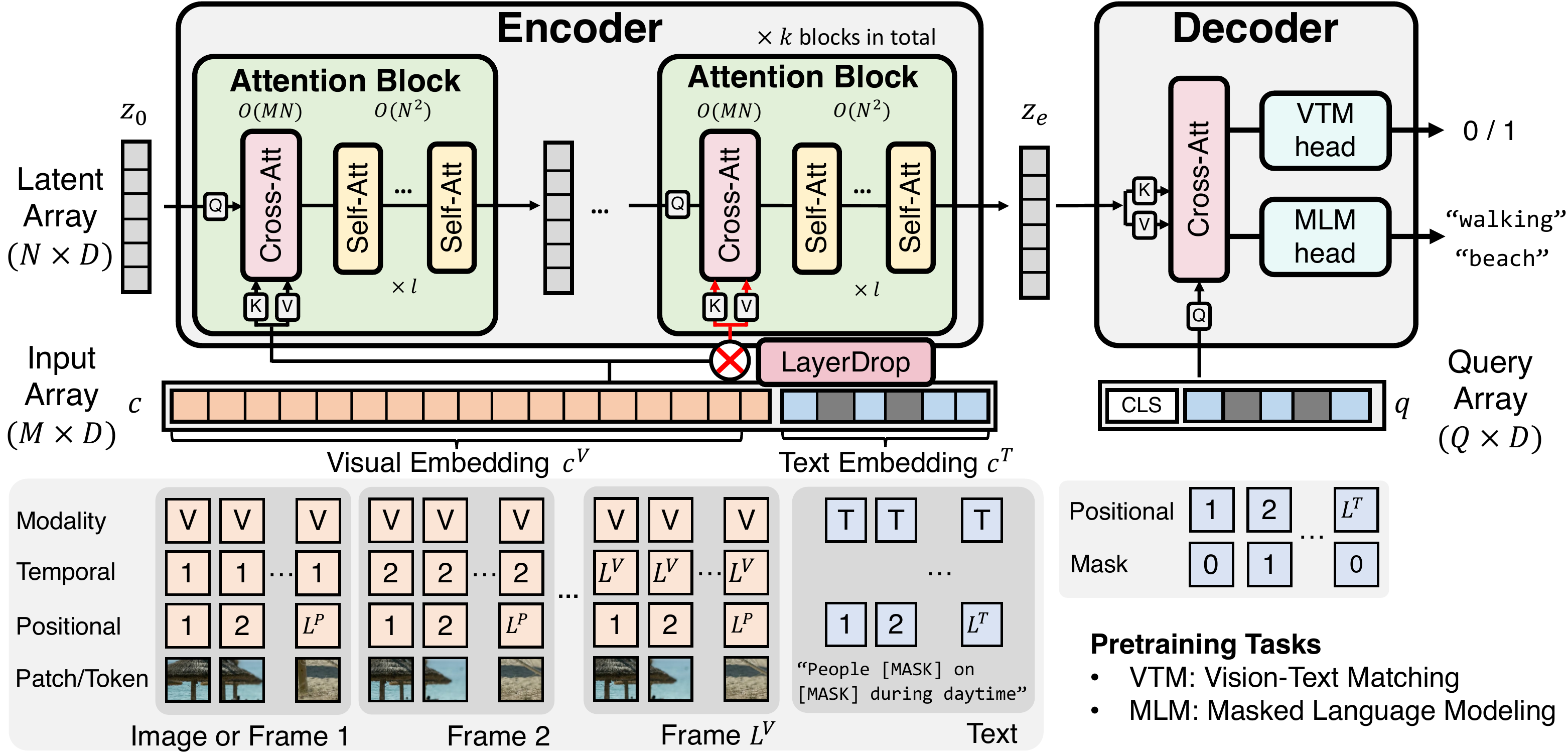}
  \caption{
  \methodname{} architecture for efficient \vl{} pretraining.
  The encoder maps the input array $c$ of length $M$ (Sec.~\ref{sec:embedding}) to the latent array $z$ of length $N$ via iterative cross-attentions (Sec.~\ref{sec:encoding}).
  Since the latent arrays are smaller than input arrays for typical \vl{} data ($N \ll M$),
  our cross-attention based encoding has higher efficiency than standard self-attention based encoding for \vl{} tasks.
  In addition, we also study dropping cross-attentions to improve latency on demand via reducing model depth (Sec.~\ref{sec:layerdrop}).
  The decoder performs structured prediction based on a cross-attention with the latent encoding $z_e$ and task-specific query array $q$ (Sec.~\ref{sec:decoding}).
}
\label{fig:perceiver_ende}
\end{figure*}

\section{Perceiver-VL}
\label{sec:method}

\methodname{} architecture consists of an input array, a latent array, and an encoder-decoder network. In the following, we explain the details of each component and how \methodname{} processes high-dimensional \vl{} data efficiently.
In Fig.~\ref{fig:perceiver_ende}, we illustrate the \methodname{} architecture.

\subsection{Vision and Language Embedding}
\label{sec:embedding}

We extend the single-modal input array of \pcv{} to the \vl{} domain, creating the input array $c$ as a concatenation of visual and text embeddings.
The embeddings are created as the sum of
(1) modality embedding,
(2) temporal embedding,
(3) positional embedding, and
(4) patch/token embedding.
Modality embedding is a learned embedding of a binary modality indicator  $\in \{\texttt{V}, \texttt{T}\}$.
Temporal embedding is a learned embedding of input video frames $\in \{1 \cdots L^V\}$, where $L^V$ is the frame length.
Note that temporal embedding is only used for videos; we do not use temporal embedding for images or text. 
Positional embedding is a learned embedding of 2D patches $\in \{1 \cdots L^P\}$ for image/video or token indices for text $\in \{1 \cdots L^T\}$, where $L^P$ is the number of patches for images, and $L^T$ is the number of text tokens.
Patch embedding is learned with a linear projection of non-overlapping image/video input patches (\eg, $32\times32$ pixels) \cite{dosovitskiy2020image}.
We treat an image as a single-frame video so that our model can flexibly process image and video input with a single architecture~\cite{bain2021frozen}.
Token embedding is a learned embedding of text tokens. 

\subsection{Iterative Mapping to Low-Dim Latent Space}
\label{sec:encoding}

Following \cite{jaegle2021perceiver1},
\methodname{} tames the quadratic computation complexity of self-attentions over high-dimensional inputs, by introducing a latent array $z$ of size $N$ (see `Latent Array' in Fig.~\ref{fig:perceiver_ende}) that aggregates information from an input array $c$ of size $M$ via iterative cross-attentions (see `Cross-Att' in Fig.~\ref{fig:perceiver_ende}).
\methodname{} encoder consists of $k$ attention blocks, each consisting of a cross-attention and $l$ self-attentions over a latent array $z$, resulting in the computational complexity of $\mathcal{O}(kMN+klN^2)$.
In comparison, a standard transformer encoder with the same number of self-attention modules has a computational complexity of $\mathcal{O}(klM^2)$.
Since in \vl{} tasks where the input size $M$ is larger than the latent array size $N$,
the change from quadratic to linear computational complexity w.r.t. $M$ can greatly increase efficiency (see Fig.~\ref{fig:plot_fi}).
To disambiguate the latent dimensions, we add a position embedding to the latent array $z$.
We add the learned positional embedding \cite{gehring2017convolutional,vaswani2017attention} 
for each latent dimension.
The choice of learned position encoding is based on simplicity; different from the findings from the single-modality experiments of \cite{jaegle2021perceiver1}, we did not find the gain from using Fourier feature position encodings \cite{stanley2007compositional,mildenhall2020nerf,jaegle2021perceiver1}, as shown in the
appendix.

\subsection{LayerDrop on Cross-Attention for Reducing Depth on Demand}
\label{sec:layerdrop}

It is the cross-attention layers that take the highest computation in the attention blocks. 
Therefore, to further improve the efficiency of \methodname{}, we apply LayerDrop~\cite{fan2019reducing} to cross-attention layers,
which allows users to control the latency by changing the number of cross-attention layers during inference.
Concretely, we apply dropout~\cite{fan2019reducing} to each cross-attention layer with probability $p^{LD}$ during pretraining (see `LayerDrop' in Fig.~\ref{fig:perceiver_ende}).
Note that we do not apply LayerDrop to the first cross-attention layer, to ensure that the model always receives the signal from input.
We study the effect of different $p^{LD}$ and the effect of varying the number of cross-attention layers during inference
(see \cref{sec:ablation_layerdrop} for details). 

\subsection{Structured Decoding with Cross-Attention and Query Array}
\label{sec:decoding}

To adapt \methodname{} to different \vl{} tasks with structured output space,
we give a query array $q$ of arbitrary length $Q$ (see `Query Array' in Fig.~\ref{fig:perceiver_ende}),
to decoder cross-attention 
and apply a task-specific head (a fully-connected layer) to the cross-attention output.
We use a decoder with a single cross-attention~\cite{jaegle2021perceiver2}.
For multi-task learning, we simply concatenate the query array for different tasks.
In the following, we describe decoder queries for two \vl{} pretraining objectives.
See appendix for the decoding details for downstream tasks.

\subsubsection{Vision-and-Language Pretraining}
We use two popular objectives in \vl{} domain for \methodname{} pretraining: Vision-Text Matching (VTM) and Masked Language Modeling (MLM).
To create the final query for the VTM and MLM tasks, we concatenate the queries for the two tasks, as illustrated in Fig.~\ref{fig:perceiver_ende}.

\paragraph{Vision-Text Matching (VTM)}
asks a model to distinguish whether a given pair of visual input (image or video) and text input matches or not.
We create an unmatched pair by replacing its visual input with a randomly selected negative one, with $50\%$ probability.
We create the VTM query with a learnable embedding ($Q=1$), illustrated as \texttt{[CLS]} in
Fig.~\ref{fig:perceiver_ende}.
We apply a linear VTM head to the corresponding
decoder output and perform binary classification.

\paragraph{Masked Language Modeling (MLM)}
asks a model to infer masked text inputs in a given context.
Following \cite{devlin2018bert}, we randomly mask 15\% of the input text tokens.
We create the MLM query by 
adding a positional embedding and a mask embedding ($Q=L^T$).
The mask embedding is a learned embedding of a binary indicator variable, where 1 indicates the masked text token.
Note that we do not feed the token embeddings to the decoder, \ie, we do not provide the text input.
In doing so, we encourage the encoder output $z_e$ to have a compact representation that contains enough information for MLM.
We apply a linear MLM head to the corresponding decoder output and use cross-entropy loss.

\begin{figure*}[t]
  \centering
\includegraphics[width=0.85\textwidth]{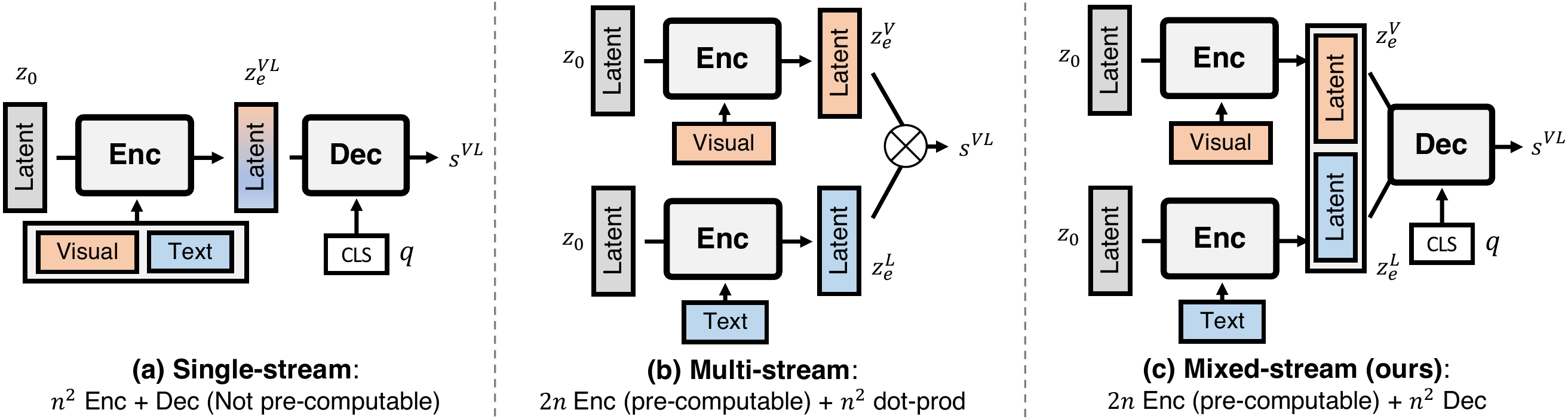}
  \caption{
  \methodname{} architectural variants for retrieval task (Sec.~\ref{sec:retrieval_arch}).
  \textbf{\textit{(a)} single-stream}: encoder jointly processes concatenated multi-modal inputs, followed by decoder.
  \textbf{\textit{(b)} multi-stream}: encoder separately processes single-modal inputs, followed by dot products.
  \textbf{\textit{(c)} mixed-stream (ours)}: encoder separately processes single-modal inputs, followed by decoder processing the concatenation of encodings.
  We included the computation complexity of each architecture under its title.
  Note that most computation happens in the encoder (\textit{Enc}).
  Because single-stream encoder does not allow pre-computation and also requires the largest computation,
  it achieves the lowest efficiency during inference, while multi-stream and mixed-stream architectures have similar efficiency (see \cref{sec:result_retrieval_architectures}).
  }
  \label{fig:perceiver_retr}
\end{figure*}

\subsection{Mixed-Stream Architecture for Cross-Modal Retrieval}
\label{sec:retrieval_arch}

In Fig.~\ref{fig:perceiver_retr}, we show two widely used architectures
used in cross-modal retrieval tasks:
(a) single-stream \cite{devlin2018bert,kim2021vilt} and (b) multi-stream \cite{radford2021learning,bain2021frozen}.
The single-stream architecture computes the multi-modal similarity score $s^{VL}$ with multiple layers of encoder,
whereas the multi-stream encoder computes the multi-modal similarity simply with a dot product between single-modality encodings $z^V_e, z^L_e$ from separate encoders.
In many real-world applications, multi-stream architectures are widely used for retrieval for their high efficiency.
This is because multi-stream architectures allow us to cache pre-computed visual encodings $z^V_e$ and simply compute dot products with text query $z^L_e$ during inference.
In contrast, single-stream architectures tend to achieve higher accuracy but require expensive computation, where a joint input array goes through multiple encoder layers.
We propose to use a `mixed-stream' architecture (Fig.~\ref{fig:perceiver_retr} (c)) that takes the best of both worlds.
Note that a similar idea has been proposed for text retrieval~\cite{humeau2019poly}.
As shown in Fig.~\ref{fig:plot_ms}, our mixed-stream architecture achieves a good accuracy-latency tradeoff.

\section{Experiment Setup}
\label{sec:expsetup}

We pretrain \methodname{} on a combination of video-text and image-text datasets, then finetune it on a set of downstream benchmarks for evaluation. Below, we explain the details of our training and evaluation setup.

\subsection{Architecture Details}

\paragraph{Model Details.}

For the \methodname{} encoder, we use $k=3$ blocks of 1 cross-attention and $l=3$ self-attentions, totaling 3 cross-attention layers and 12 self-attention layers. The decoder has 1 cross-attention layer. We follow BERT$_{\text{BASE}}$~\cite{devlin2018bert} and ViT-B/32 \cite{dosovitskiy2020image} to use a hidden size of 768 and 12 attention heads. We follow ViT-B/32 \cite{dosovitskiy2020image} to use image (and video frame) size 384 and patch size 32. We use PyTorch \cite{NEURIPS2019_9015} to implement our model in experiments.

\paragraph{LayerDrop on Cross-Attention.}

We set a probability of $p^{LD}=0.5$ to apply dropout to the cross-attention layers during \vl{} pretraining.
Note that we do not apply dropout to the first cross-attention, to ensure that input signal always goes into the latent array.
We analyze the effect of using LayerDrop during pretraining, finetuning, and inference, as shown in Table~\ref{tab:layerdrop_msrvtt}.

\begin{figure*}[t]
  \centering
\includegraphics[width=0.8\textwidth]{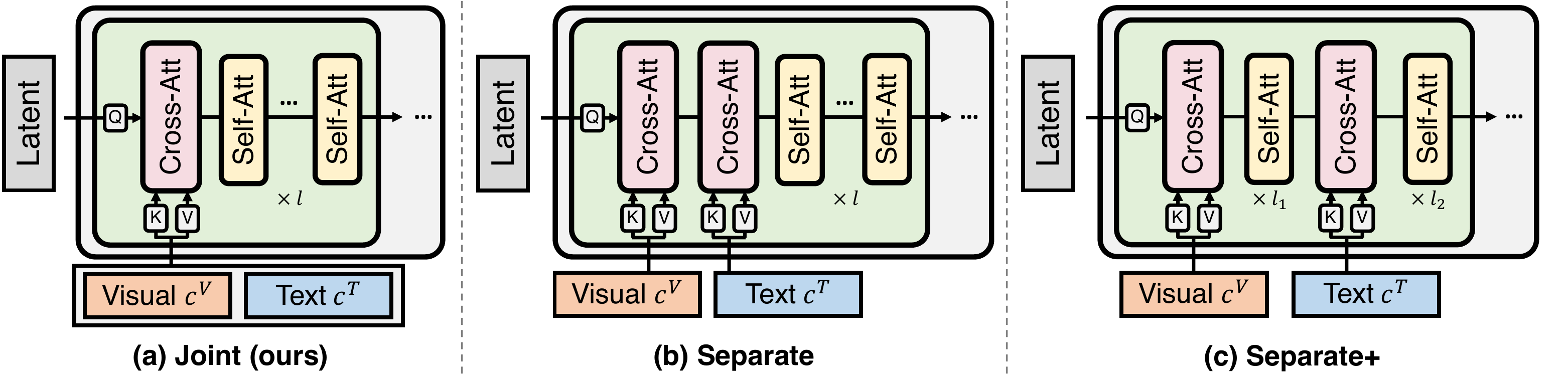}
  \caption{
  Different multi-modal encoding schemes.
  \textbf{\textit{(a)} Joint}: Embeddings of each modality are concatenated and jointly encoded with a single cross-attention to the latent space.
  \textbf{\textit{(b)} Separate}: using separate cross-attention for mapping each modality to the latent space.
  \textbf{\textit{(c)} Separate+}: using self-attention after each single-modal cross-attention.
  }
  \label{fig:modality_aggre}
\end{figure*}

\paragraph{Modality Aggregation.} 

By default, we map the multi-modal inputs to the latent space by creating an input array based on the concatenation of visual and textual inputs (namely \textit{Joint} encoding).
We also explore two other ways of combining the two modalities:
encoding each modality serially with separate cross-attentions, then applying self-attentions (\textit{Separate} encoding);
encoding each modality serially with separate cross-attentions with self-attentions between them (\textit{Separate+} encoding).
We illustrate these aggregation strategies in Fig.~\ref{fig:modality_aggre}.
In our ablation study, we found that the three methods perform comparably, where the \textit{Joint} encoding has the least computation. Therefore, we adopt the \textit{Joint} encoding as our default modality aggregation method. See appendix for the detailed experiments.

\subsection{Weight Initialization from Vision Transformers}
\label{sec:weightinit}

To compare with recent methods that use pretrained visual backbone models, we experiment with initializing weights of \methodname{} with two popular models: ViT-B/32~\cite{dosovitskiy2020image} and CLIP (ViT-B/16)~\cite{radford2021learning}.
As these models have 12 self-attention layers, we insert 3 cross-attention layers after every 4 self-attention layers (before 1st/5th/9th).

\paragraph{Two-stage training.}
Since transformer models do not have cross-attention layers, the cross-attention weights could not \textit{warm-start}. To stabilize training, after initializing the \methodname{} weights from CLIP parameters, we first train only the cross-attention layers, while freezing all other modules. After initial convergence (\eg, 1 epoch in \msrvtt{}~\cite{xu2016msr}), we train the whole model jointly.
In our experiment, this two-stage training strategy achieves better weight transfer than single-stage training (see appendix).

\subsection{Pretraining Setups}
\label{sec:pretraining_setup}

\paragraph{Video-Text and Image-Text Datasets.}

We follow \frozen{} \cite{bain2021frozen} to pretrain \methodname{} on both video-text and image-text data. 
For image-text dataset, we use Conceptual Captions (CC) \cite{sharma2018conceptual}, which consists of 3M image-text pairs.
For video-text dataset, we use \webvid{} \cite{bain2021frozen}, which consists of 2.5M video-text pairs, with an average video length of 18 seconds.

\paragraph{Training Details.}
We use Adam optimizer~\cite{kingma2014adam} with a learning rate 1e-5 and weight decay 0.001. We set 200k training steps for joint Video-Text \& Image-Text pretraining.
We use the batch size of 4096 with gradient accumulation on 4 RTX 2080Ti GPUs for 14 days.

\subsection{Downstream Tasks}
\label{sec:downstream_setup}

After pretraining, we evaluate \methodname{} on various \vl{} benchmarks,
covering cross-modal retrieval and visual question answering for both video-text and image-text datasets.

\subsubsection{Video-Text Tasks}

For video retrieval, we use \msrvtt{}~\cite{xu2016msr}, \lsmdc{}~\cite{rohrbach2015dataset}, \didemo{}~\cite{anne2017localizing}, \activitynet{} Captions~\cite{krishna2017dense}.
For video question answering, we use \tgifqa{} \cite{jang2017tgif} and \msrvttqa{}.

\paragraph{Dataset Details.}

\msrvtt{} contains 10K web video clips, with 20 captions for each clip. 
\lsmdc{} contains 118,081 short clips from 202 movies with each clip containing one caption. 
\didemo{} contains 10,464 videos with 40,543 temporally localized sentences.
\activitynet{} Captions have 20k videos with 3.65 temporally localized sentences per video on average, resulting in 100k sentences in total. Videos have an average duration of 180 seconds.
For \didemo{} and \activitynet{} Captions, we follow previous work~\cite{liu2019use,lei2021less} to use paragraph-to-video retrieval, where we concatenate all sentences from the same video as a single text query for retrieval.
\tgifqa{} contains 165K QA pairs from 72K animated GIF videos. We follow \cite{lei2021less} to evaluate our model on three \tgifqa{} tasks: action, transition, and frame.
\msrvttqa{} contains 10K videos with 243K open-ended questions collected from \msrvtt{} videos.

\paragraph{Training Details.}
We use Adam optimizer~\cite{kingma2014adam} with a learning rate 1e-5 and weight decay $0.001$.
We use 16 frames for \activitynet{} Captions and 8 frames for other tasks.
We use frame size $384\times384$, and a maximum text length 40 for all tasks.

\subsubsection{Image-Text Tasks}

For image retrieval, we use \flickrthirty{} \cite{plummer2015flickr30k}.
For visual question answering, we use \vqa{} \cite{antol2015vqa} and \nlvr{} \cite{suhr2018corpus}.

\paragraph{Dataset Details.}
\vqa{} contains 204,721 images from \coco{} \cite{lin2014microsoft}, with a minimum of 3 questions per image and 10 grounded answers. \nlvr{} contains 107,292 examples of sentences grounded in pair of images. \flickrthirty{} dataset has 31,000 images collected from \flickr{} each with 5 sentences.

\paragraph{Training Details.}
We use Adam optimizer with a learning rate of 1e-4 and weight decay of 0.001.
We use image size $384\times384$ and a maximum text length 40 for all tasks.

\begin{table*}[t]
\begin{center}

\setlength{\tabcolsep}{4pt}
\resizebox{.93\textwidth}{!}{
\begin{tabular}{lllcccccccc}
\toprule
\multirow{2}{*}{Model} & 
\multirow{2}{*}{Pretraining Datasets} & \multirow{2}{*}{Visual Backbone} & 
\multicolumn{4}{c}{Retrieval $\uparrow$} &
\multicolumn{2}{c}{QA Acc. $\uparrow$} &
\multirow{2}{*}{GFLOPs $\downarrow$} &
\multirow{2}{*}{Time (ms) $\downarrow$}
\\
\cmidrule(lr){4-7}  \cmidrule(lr){8-9}
& & & MSR & DDM & LSM & ACT & TGIF & MSR & & \\

\midrule
\multicolumn{3}{l}{\textcolor{gray}{Models using other input modalities (\eg, audio)}} \\
\textcolor{gray}{\hero{} \cite{li2020hero}} & \textcolor{gray}{TV/\htm{}} 
& \textcolor{gray}{ResNet152+Slowfast \cite{he2016deep,feichtenhofer2019slowfast}}& \textcolor{gray}{20.5} & \textcolor{gray}{-} & \textcolor{gray}{-} & \textcolor{gray}{-} & \textcolor{gray}{-} & \textcolor{gray}{-} & \textcolor{gray}{935.2} & \textcolor{gray}{2200.0} \\
\textcolor{gray}{\mmt{} \cite{gabeur2020multi}} & \textcolor{gray}{\htm{} }
& \textcolor{gray}{S3D+VGG+DenseNet161 \cite{xie2018rethinking,hershey2017cnn,huang2017densely}} & \textcolor{gray}{26.6} & \textcolor{gray}{-} & \textcolor{gray}{12.9} & \textcolor{gray}{-} & \textcolor{gray}{-} &  \textcolor{gray}{-} & \textcolor{gray}{-} & \textcolor{gray}{-}\\
\textcolor{gray}{\alvnet{} \cite{rouditchenko2020avlnet}} & \textcolor{gray}{\htm{} }
& \textcolor{gray}{ResNet152+ResNeXt \cite{he2016deep,xie2017aggregated}} & \textcolor{gray}{27.1} & \textcolor{gray}{-} & \textcolor{gray}{17.0} & \textcolor{gray}{-} & \textcolor{gray}{-} & \textcolor{gray}{-} & \textcolor{gray}{153.4} & \textcolor{gray}{2000.0}\\
\midrule

\multicolumn{3}{l}{\textcolor{gray}{Models with CLIP initialization}} \\
\textcolor{gray}{Hunyuan(+DSL) \cite{min2022hunyuan_tvr}} & -
& \textcolor{gray}{CLIP (ViT-B/16)} & \textcolor{gray}{\textbf{55.0}} & \textcolor{gray}{\textbf{52.1}} & \textcolor{gray}{\textbf{29.7}} & \textcolor{gray}{\textbf{57.3}} & \textcolor{gray}{-} & \textcolor{gray}{-} & \textcolor{gray}{2022.8} & \textcolor{gray}{-}
\\
\textcolor{gray}{CLIP2TV \cite{gao2021clip2tv}} & \textcolor{gray}{-}
& \textcolor{gray}{CLIP (ViT-B/16)} & \textcolor{gray}{49.3} & \textcolor{gray}{45.5} & \textcolor{gray}{-} & \textcolor{gray}{44.1} & \textcolor{gray}{-} & \textcolor{gray}{-} & \textcolor{gray}{2212.3} & \textcolor{gray}{-}
\\
\textcolor{gray}{DRL \cite{wang2022disentangled}} & \textcolor{gray}{-}
& \textcolor{gray}{CLIP (ViT-B/32)} & \textcolor{gray}{48.8} & \textcolor{gray}{49.0} & \textcolor{gray}{26.5} & \textcolor{gray}{46.2} & \textcolor{gray}{-} & \textcolor{gray}{-} & \textcolor{gray}{511.0} & \textcolor{gray}{320.0}
\\
\textcolor{gray}{CAMoE(+DSL) \cite{cheng2021improving}} & \textcolor{gray}{-}
& \textcolor{gray}{CLIP (ViT-B/32)} & \textcolor{gray}{47.3} & \textcolor{gray}{-} & \textcolor{gray}{25.9} & \textcolor{gray}{-} & \textcolor{gray}{-} & \textcolor{gray}{-} & \textcolor{gray}{399.7} & \textcolor{gray}{-}
\\
\textcolor{gray}{MDMMT-2 \cite{kunitsyn2022mdmmt}} & \textcolor{gray}{-}
& \textcolor{gray}{CLIP (ViT-B/32)} & \textcolor{gray}{48.5} & \textcolor{gray}{-} & \textcolor{gray}{26.9} & \textcolor{gray}{-} & \textcolor{gray}{-} & \textcolor{gray}{-} & \textcolor{gray}{-} & \textcolor{gray}{-}
\\
\textcolor{gray}{Ours$^{N=32, \text{Multi}}$} & \textcolor{gray}{\cc{}/\webvid{}}
& \textcolor{gray}{CLIP (ViT-B/16)} & \textcolor{gray}{45.9} & \textcolor{gray}{-} & \textcolor{gray}{-} & \textcolor{gray}{-} & \textcolor{gray}{-} & \textcolor{gray}{-} & \textcolor{gray}{\textbf{80.0}} & \textcolor{gray}{\textbf{80.0}} \\

\midrule
\htm{} \cite{miech2019howto100m} & \htm{} 
& ResNet152+ResNeXt \cite{he2016deep,xie2017aggregated} & 14.9  & - & 7.1 & - & - & - & 164.3 & 1100.0\\
\clipbert{} \cite{lei2021less} & \coco{}/\cc{}
& ResNet50 \cite{Jiang2020} &  22.0 &  20.4 & - & 21.3 & 60.3 & 37.4 & 340.0 & 700.0 \\
\frozen{} \cite{bain2021frozen} & \cc{}/\webvid{}
& Timesformer-B/16 \cite{bertasius2021space} & 31.0 & \textbf{31.0} &  15.0 & - & - & - & 89.0 & 260.0 \\
Ours$^{N=128, \text{Mixed}}$ & \cc{}/\webvid{}
& ViT-B/32 \cite{dosovitskiy2020image} & \textbf{32.6} & 30.5 & \textbf{15.8} & \textbf{33.9} & \textbf{69.2} & \textbf{43.2} & \textbf{43.9} & \textbf{72.0}\\

\bottomrule
\end{tabular}
}
\end{center}
\caption{
Finetuning performance on text-to-video retrieval and video question answering benchmarks.
We report R@1 for text-to-video retrieval tasks (see appendix for R@5/R@10) and report QA accuracy on the FrameQA task.
\textit{GFLOPs} shows the inference cost on a single sample, and \textit{Time (ms)} indicates the average inference time across all samples on \msrvtt{} val split.
For a fair comparison, we gray out 1) the models that use input modalities other than video and text (\eg, audio) and 2) the models that use CLIP visual encoder \cite{radford2021learning} (the cross-attention layers of \methodname{} cannot be initialized with CLIP parameters and trained from scratch; see the discussion in \cref{sec:overall_comparison}).
\textit{MSR}=\msrvtt{}, \textit{DDM}=\didemo{}, \textit{LSM}=\lsmdc{}, \textit{ACT}=\activitynet{}, \textit{TGIF}=\tgifqa{}.
$^{N=128}$ means latent size N=128.
$^{\text{Multi}}$ and $^{\text{Mixed}}$ mean multi-stream and mixed-stream respectively.
}
\label{tab:downstream_videotext}
\end{table*}

\begin{table*}[t]
\setlength{\tabcolsep}{4pt}
\begin{center}
\resizebox{.93\textwidth}{!}{
\begin{tabular}{lllcccccc}
\toprule
\multirow{2}{*}{Model} & 
\multirow{2}{*}{Pretraining Datasets} & \multirow{2}{*}{Visual Backbone} & 
\multicolumn{1}{c}{Retrieval $\uparrow$} &
\multicolumn{2}{c}{QA Accuracy $\uparrow$} &
\multirow{2}{*}{GFLOPs $\downarrow$} & 
\multirow{2}{*}{Time (ms) $\downarrow$}
\\
\cmidrule(lr){4-4}  \cmidrule(lr){5-6} 
& & & \flickrthirty{} & \vqa{} & \nlvr{} & \\

\midrule

\multicolumn{3}{l}{\textcolor{gray}{Models using additional object tag inputs}} \\
\textcolor{gray}{\vinvl{} \cite{zhang2021vinvl}} & \textcolor{gray}{\coco{}/\cc{}/\sbu{}/\flickr{}/\oi{}*} & \textcolor{gray}{Faster-RCNN \cite{zhang2021vinvl}} & \textcolor{gray}{-} & \textcolor{gray}{75.95} & \textcolor{gray}{83.08}  & \textcolor{gray}{1023.3} & \textcolor{gray}{800.0} \\
\textcolor{gray}{\oscar{} \cite{li2020oscar}} & \textcolor{gray}{\coco{}/\cc{}/\sbu{}/\flickr{}*}  & \textcolor{gray}{Faster-RCNN \cite{anderson2018bottom}} & \textcolor{gray}{-} & \textcolor{gray}{73.16} & \textcolor{gray}{78.36} & \textcolor{gray}{956.4} & \textcolor{gray}{1000.0} \\
\midrule

\uniter{} \cite{chen2019uniter} & \coco{}/\cc{}/\sbu{}/\vg{} & Faster-RCNN \cite{anderson2018bottom} & \textbf{72.5} & \textbf{72.70} & 75.80 & 949.9 & 1000.0 \\
\vilt{} \cite{kim2021vilt} & \coco{}/\cc{}/\sbu{}/\vg{} & ViT-B/32 \cite{dosovitskiy2020image} & 64.4 & 71.26 & \textbf{76.13} & 55.9 & 32.0 \\
Ours$^{N=128}$ & \coco{}/\cc{}/\sbu{}/\vg{} & ViT-B/32 \cite{dosovitskiy2020image} & 62.4 & 71.62 & 75.53 & \textbf{30.5} & \textbf{18.0} \\

\midrule

\lxmert{} \cite{tan2019lxmert} & \coco{}/\vg{}* & Faster-RCNN \cite{anderson2018bottom} &  - & \textbf{72.42} & 74.50  & 952.0 & 1100.0 \\
\visualbert{} \cite{li2019visualbert} & \coco{} & Faster-RCNN \cite{anderson2018bottom} & - & 70.80 & 67.00  & 425.0 & 1000.0 \\
\pixelbert{} \cite{huang2020pixel} & \coco{}/\vg{} & ResNet50 \cite{he2016deep} & 53.4 &  71.35 & 72.40 & 136.8 & 150.0 \\

Ours$^{N=128}$ & \coco{}/\vg{} & ViT-B/32 \cite{dosovitskiy2020image} & \textbf{61.7} & 70.45 & \textbf{74.87} & \textbf{30.5} & \textbf{18.0} \\

\midrule
\frozen{} \cite{bain2021frozen} & \cc{}/\webvid{} &  Timesformer-B/16 \cite{bertasius2021space}  & 61.0 & - & -  & 63.9 & 70.0 \\

Ours$^{N=64}$ & \cc{}/\webvid{} & ViT-B/32 \cite{dosovitskiy2020image} & 61.0 & 70.12 & 74.52 & \textbf{17.0} & \textbf{8.0} \\
Ours$^{N=128}$ & \cc{}/\webvid{} & ViT-B/32 \cite{dosovitskiy2020image} & \textbf{61.8} & \textbf{70.91} & \textbf{75.44} & 30.5 & 18.0 \\
\bottomrule
\end{tabular} 
}
\end{center}
\caption{
Finetuning performance on text-to-image retrieval and visual question answering benchmarks.
For \nlvr{}, we show Test-P accuracy. For \flickrthirty{}, we show text-to-image retrieval R@1 (see appendix for R@5/R@10). Note that for brevity, we only show the image or video source datasets for \textit{Pretraining Datasets}; the datasets that added additional text annotations are not included in the column (we use * to highlight them). For example, \lxmert{} is trained with image-text datasets COCO and VG, as well as the three QA datasets based on COCO and VG images, \ie, \vqa{}, \vgqa{} and \gqa{}.
We also gray out models that use additional object tags in the first block and are not comparable to our model.
\textit{GFLOPs} shows the inference cost on a single sample, \textit{Time (ms)} indicates the average inference time over all samples in \vqa{} minival split.
For a fair comparison, we gray out models that are pretrained with more data. $^{N=128}$ means latent size N=128.
}
\label{tab:downstream_imagetext}
\end{table*}

\section{Results and Analysis}
\label{sec:results}

We first compare \methodname{} with recent methods in video-text / image-text benchmarks, where it achieves the highest efficiency while maintaining competitive performance (Sec.~\ref{sec:overall_comparison}).
Then we analyze the efficiency of \methodname{} in detail (Sec.~\ref{sec:efficiency}).
In appendix,
we also present ablation studies of different architectural components and training strategies for \methodname{}.

\subsection{Comparison to State-of-the-Art}
\label{sec:overall_comparison}

In Table \ref{tab:downstream_videotext},
we compare \methodname{} with the state-of-the-art video-text models on 
4 text-to-video retrieval (MSRVTT, DiDeMo, LSMDC, ActivityNet) and 2 video question answering (TGIF, MSRVTT-QA) benchmarks. 
In Table \ref{tab:downstream_imagetext},
we compare \methodname{} with state-of-the-art image-text models on text-to-image retrieval (\flickrthirty{}) and 2 visual question answering (\vqa{}, \nlvr{}) benchmarks.
The closest baseline of our model is \frozen{}~\cite{bain2021frozen}, as it is pretrained on the same pretraining dataset (\cc{}/\webvid{}) and handles both images and videos in a single architecture.
\methodname{} achieves competitive performance across the board, for both image-based and video-based tasks, while maintaining significantly higher efficiency. \methodname{} has the lowest GFLOPs and inference time (see the rightmost columns of the tables). 

As some recent video retrieval models adopt CLIP \cite{radford2021learning} trained on 400M image-text pairs from the Web, we also provide experiments using CLIP checkpoint.
Concretely, we use the self-attention layers of the CLIP visual encoder to handle compact latent spaces and insert cross-attentions with randomly initialized weights.
The use of CLIP significantly improves the retrieval performance (\eg, 32.6$\rightarrow$45.9 on \msrvtt{} R@1).
However, there is a certain gap between our model and the baselines, because CLIP self-attention layers were trained to handle image patches, rather than compact latent spaces.
Thus, we gray out the CLIP-based results in Table \ref{tab:downstream_videotext} to highlight the fact that our models are not directly comparable to transformer-based models.
We expect that a better weight initialization (\eg, from a \pcv{} architecture trained on 400M image-text pairs) would further improve the performance of our models.

\subsection{Efficiency Analysis}
\label{sec:efficiency}

\begin{figure*}[t]
  \centering
  \includegraphics[width=0.36\textwidth]{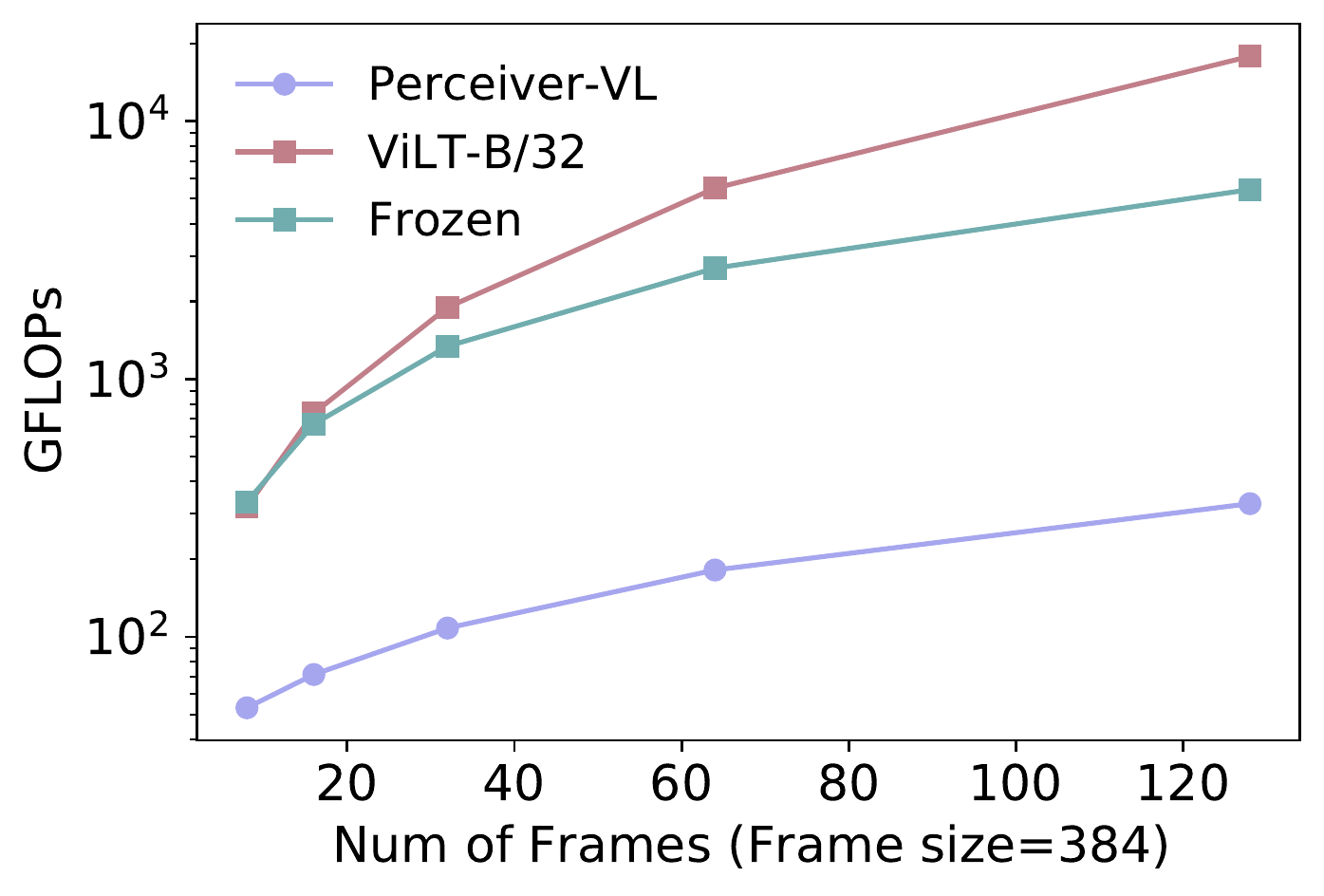}
  \includegraphics[width=0.36\textwidth]{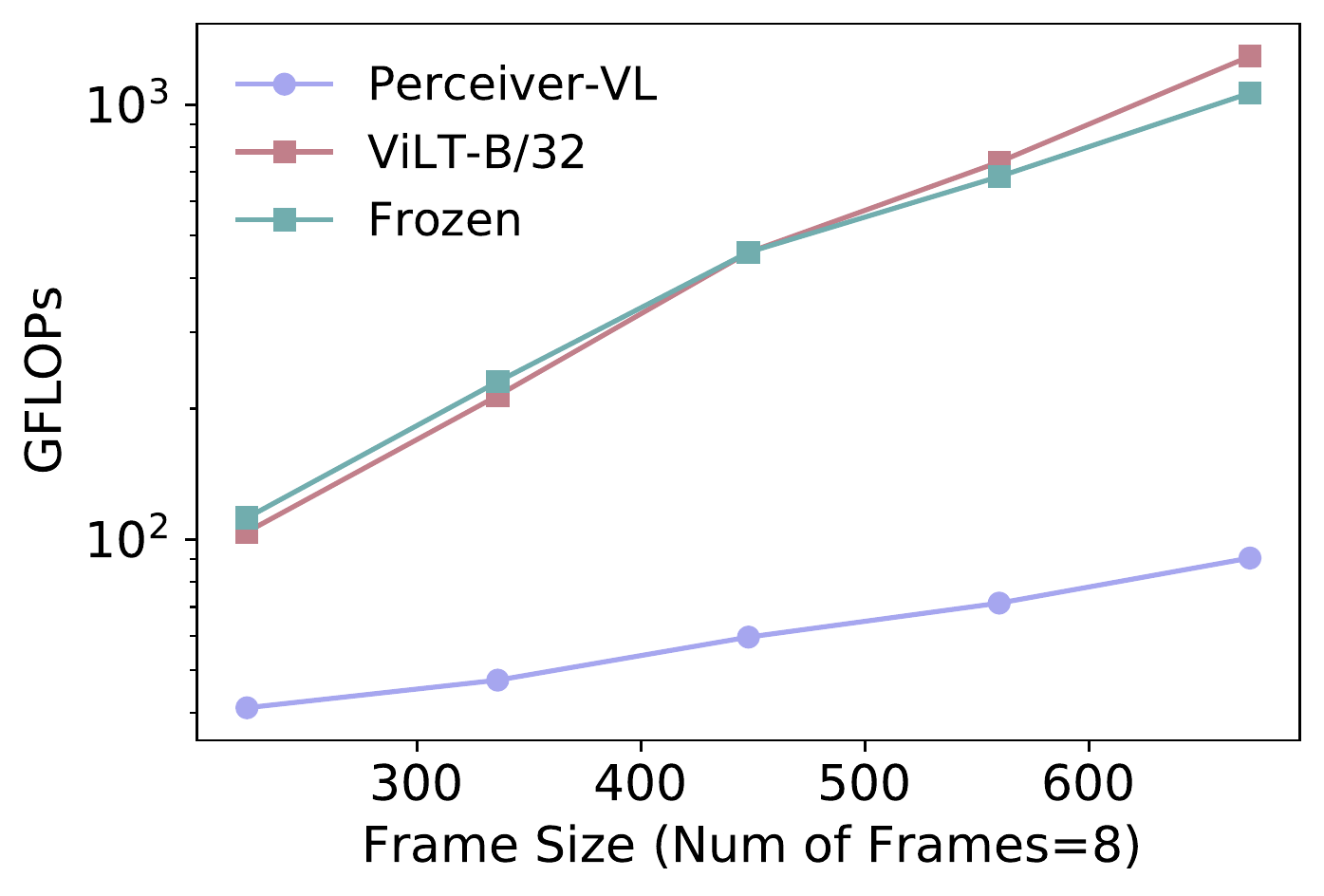}
  \caption{
  Input scaling comparison between \methodname{}, \vilt{}~\cite{kim2021vilt}, and \frozen{}~\cite{bain2021frozen} for video inputs with different the number of frames (\textit{left}) and frame size (\textit{right}).
  Note that the GFLOPs are illustrated in log-scale.
  On both plots, \methodname{} shows remarkably lower computation compared to the other two models.
  }
  \label{fig:plot_fi}
\vspace{-10pt}
\end{figure*}

\subsubsection{Scaling Input Array}
In Fig.~\ref{fig:plot_fi}, we compare the computations of \methodname{}, \vilt{}~\cite{kim2021vilt}, and \frozen{}~\cite{bain2021frozen} for video inputs of different scales, by varying the number of frames (\textit{left}) and the frame size (\textit{right}).
All three models have 12 self-attention layers with hidden size 768.
Powered by efficient cross-attention-based encoding, \methodname{} shows remarkably better scalability (lower GFLOPs) than \vilt{} and \frozen{} in both plots.

\begin{figure}
  \centering
  \includegraphics[width=.8\linewidth]{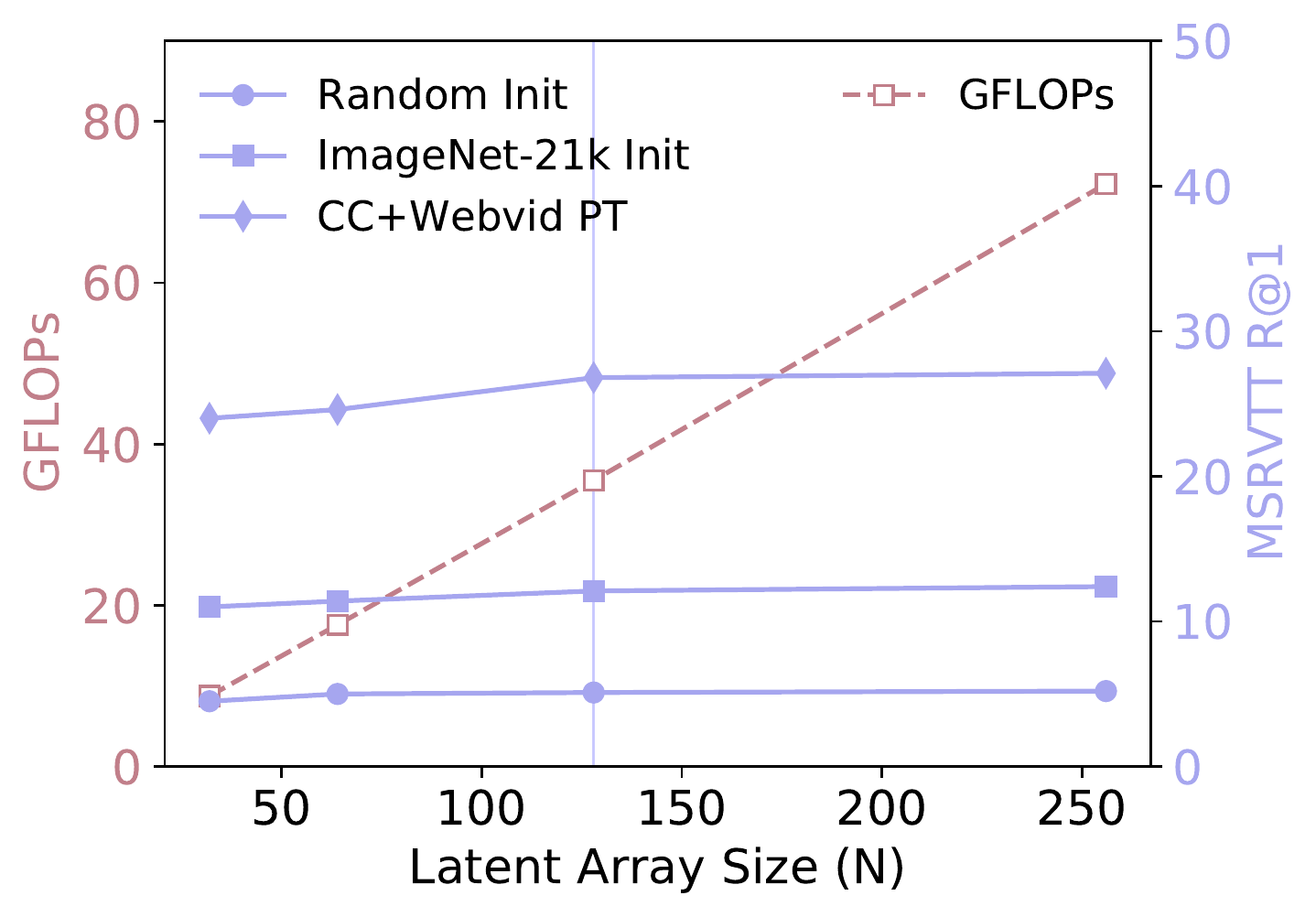}
  \caption{
  Efficiency-Accuracy tradeoff of using different latent array size $N$ during finetuning on \msrvtt{}.
  During pretraining, we use the latent array size $N=128$ (blue vertical line). We use mixed-stream architecture by default.
  }
  \label{fig:plot_fl_msrvtt}
\end{figure}

\subsubsection{Scaling Latent Array}

We study the effect of varying the latent array size $N$ to explore whether we can further improve the efficiency of \methodname{}.
In Fig.~\ref{fig:plot_fl_msrvtt}, we show the effect of varying the latent array sizes during finetuning in terms of computation and downstream performance on \msrvtt{}.
We use $N{=}128$ during pretraining. 
When scaling up or down the latent array for a pretrained model, we simply initialize a new latent array where we empirically find that it gives a similar performance compared to interpolating the pretrained latent array.
We can see that the GFLOPs scale linearly with $N$, while the retrieval performance remains reasonably well in three different pretraining setups (\eg, CC+Webvid PT: $24.0\rightarrow24.6\rightarrow26.8\rightarrow 27.1$ with latent array size: $32\rightarrow64\rightarrow128\rightarrow 256$).

\vspace{-5pt}

\begin{figure}
  \centering
  \includegraphics[width=.8\linewidth]{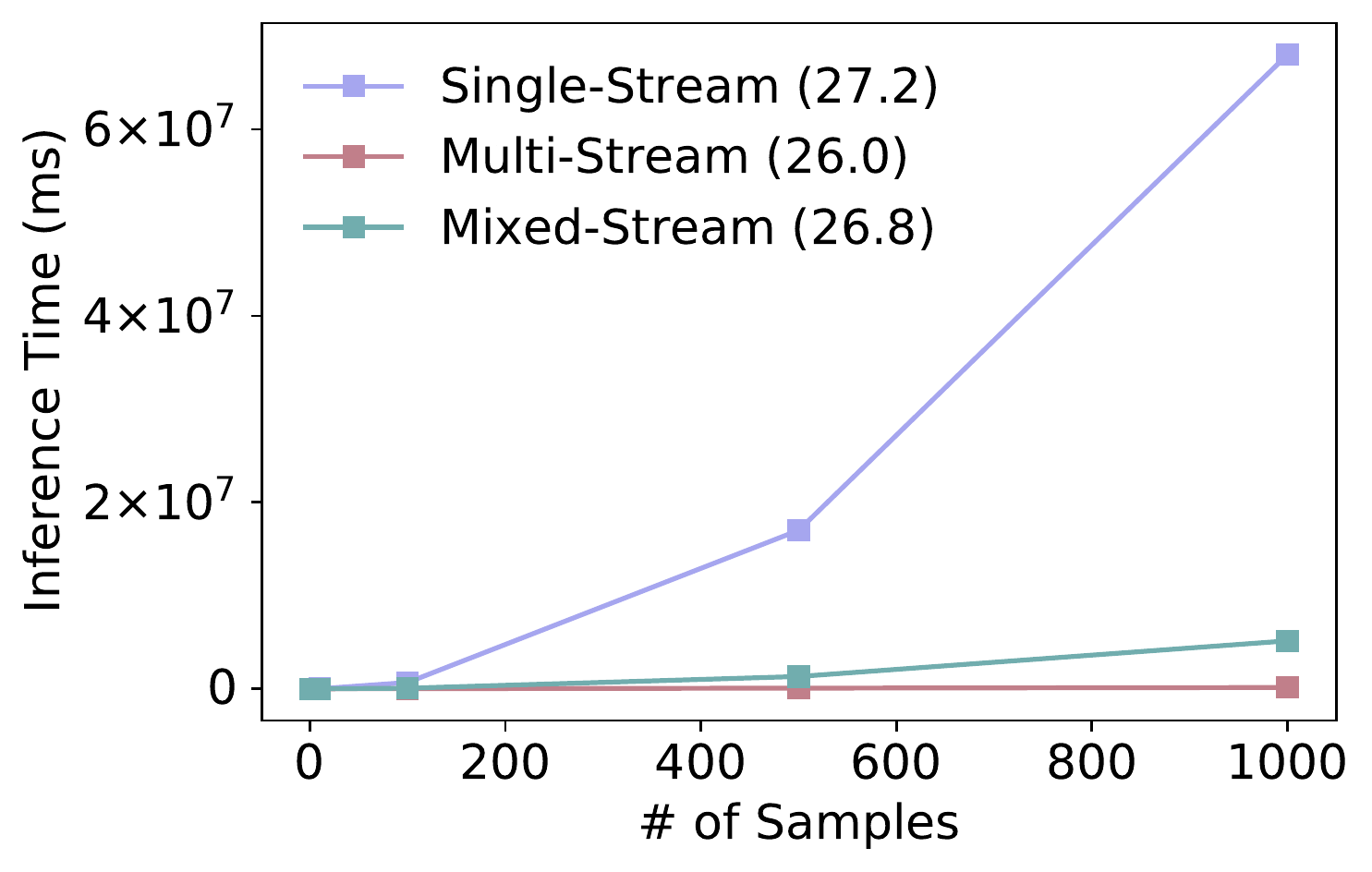}
  \caption{
  Comparison of retrieval architectures with different number of \msrvtt{} video-text pairs.
  The inference time is calculated from the total retrieval time.
  }
  \label{fig:plot_ms}
\vspace{-10pt}
\end{figure}

\subsubsection{Mixed-Stream Architecture for Retrieval}
\label{sec:result_retrieval_architectures}

In Fig.~\ref{fig:plot_ms}
we compare different retrieval architecture variants discussed in Sec.~\ref{sec:retrieval_arch} and Fig.~\ref{fig:perceiver_retr}, in terms of accuracy and inference time on \msrvtt{} val split.
The single-stream architecture achieves the highest R@1 (27.2), but also takes the longest inference time.
The multi-stream architecture achieves the lowest R@1 (26.0), with the shortest inference time.
Our mixed-stream architecture achieves a good accuracy-latency tradeoff, with R@1 (26.8) close to single-stream architecture, while running significantly faster.

\begin{table}[ht]
\setlength{\tabcolsep}{6pt}
\begin{center}
\resizebox{0.4\textwidth}{!}{
\begin{tabular}{ccc cc}
\toprule
\multicolumn{3}{c}{\# Cross-attentions in encoder} & \multirow{2}{*}{\msrvtt{} R@1} & Time\\
\cmidrule{1-3}
Pretraining & Finetuning & Inference  & & (ms) \\
\midrule
3   & 3 & 3 & 26.4 & 72.0\\
$1 \sim 3$ ($0.5$) & 3 & 3 & 26.8 & 72.0 \\
$1 \sim 3$ ($0.7$) & 3 & 3 & 26.6 & 72.0 \\
\midrule
$1 \sim 3$ ($0.5$) & 1 & 1 & 26.1 & \textbf{58.0} \\
$1 \sim 3$ ($0.5$) & 3 & 3 & 26.8 & 72.0\\
$1 \sim 3$ ($0.5$) & 3 & 1 & 24.0 & \textbf{58.0}\\
\midrule
$1 \sim 3$ ($0.5$) & $1 \sim 3$ ($0.5$) & 1 & 26.3 & \textbf{58.0}\\
$1 \sim 3$ ($0.5$) & $1 \sim 3$ ($0.5$) & 3 & 27.1 & 72.0\\
\bottomrule
\end{tabular} 
}
\end{center}
\caption{
Accuracy and inference time on \msrvtt{} retrieval with varied number of cross-attentions in \methodname{} mixed-stream encoder.
We include the layer dropout probability $p^{LD}$ in brackets if used.
Note that \methodname{} has 3 cross-attention layers in the encoder, and we do not apply dropout to the first cross-attention in the encoder ($p^{LD}=0$) to ensure that the latent array always receives signals from the input.
}
\label{tab:layerdrop_msrvtt}
\vspace{-10pt}
\end{table}

\vspace{-10pt}

\subsubsection{LayerDrop to Encoder Cross-Attentions}
\label{sec:ablation_layerdrop}

In Table~\ref{tab:layerdrop_msrvtt} we analyze the effect of applying LayerDrop (LD)~\cite{fan2019reducing} to cross-attention layers in encoder, as discussed in Sec.~\ref{sec:layerdrop} on \msrvtt{} retrieval. We use the mixed-stream architecture as the default setting.
First, we observe that LD acts as a regularizer, as we see that LD improves the \msrvtt{} accuracy in the first block, while increasing $p^{LD}$ too high $0.5\rightarrow0.7$ does not help the performance ($26.8 \rightarrow 26.6$).
The last row in the bottom block achieves the best accuracy (27.1), with LD during both pretraining and finetuning.
Second, removing cross-attention layers without LD during finetuning hurts performance (see  $26.1 \rightarrow 24.0$  in the middle block).
Lastly, with LD during finetuning, the latency of the inference time can be reduced by 19.4\% (72.0 ms $\rightarrow$ 58.0 ms), with minimal accuracy drop (see 27.1 $\rightarrow$ 26.3 in the bottom block).
This indicates that, with an LD-finetuned model, we can control its latency on demand at the inference time by varying the number of cross-attention layers, without storing checkpoints of multiple models.
\section{Conclusion}
\label{sec:conclusion}

In this work, we present \methodname{}, a \vl{} framework that efficiently handles high-dimensional multi-modal inputs such as long videos and text.
The efficiency of \methodname{} comes from linear complexity based on iterative cross-attention,
LayerDrop on cross-attention layers,
and a mixed-stream architecture for cross-modal retrieval.
Experiments on diverse \vl{} benchmarks show that our framework has a remarkably higher efficiency than state-of-the-art models while achieving competitive or better performance.
Moreover, we comprehensively analyze the efficiency of our framework, including measuring scalability in terms of input and latent array size,
reducing latency by dropping cross-attention layers, comparing architecture variants, and an ablation study on model training details.
Future works would further explore efficient VL modeling with more diverse tasks,
larger-scale pretraining of \methodname{}, and more effective knowledge transfer methods from models with heterogeneous architectures such as transformers.

\section*{Acknowledgments}
We thank the reviewers for their helpful comments.
This work was supported by ARO Award W911NF2110220, DARPA KAIROS Grant FA8750-19-2-1004, ONR Grant N000141812871, and NSF-AI Engage Institute DRL-211263. The views, opinions, and/or findings contained in this article are those of the authors and not of the funding agency.


{\small
\bibliographystyle{ieee_fullname}
\bibliography{bib}
}

{
\appendix

\section*{Appendix Overview}

In the appendix, we provide the content as follows:
\begin{itemize}
    \item Decoding for downstream tasks (\cref{sup_sec:decoding})
    \item Additional efficiency analysis on \vqa{} (\cref{sup_sec:efficiency})
    \item Ablation studies on modality aggregation, pretaining dataset, positional encoding, and two-stage training for CLIP initialization (\cref{supp_sec:ablation})
    \item Full experiment result tables including R@1/R@5/R@10 metrics (\cref{sup_sec:full_experiment_results})
\end{itemize}

\section{Decoding For Downstream Tasks}
\label{sup_sec:decoding}

We continue \cref{sec:decoding} in the main paper to discuss the decoding details for downstream tasks.

\subsection{Visual Question Answering}

We tackle visual question answering tasks as a classification task (e.g., \vqa{}), by choosing the right answer from a predefined answer vocabulary, following \cite{lu2016hierarchical}.
Similar to the VTM task, we create a decoder query with a \texttt{[CLS]} embedding ($Q=1$), then apply a classification head with cross-entropy loss.

\subsection{Cross-Modal Retrieval}

We tackle cross-modal retrieval tasks by first estimating the multi-modal similarity scores $s^{VL}$ of image-text or video-text pairs,
then retrieving content by ranking the similarity scores.
We study different types of architecture for this task and explain the details in Sec.~\ref{sec:retrieval_arch}.
For multi-stream architecture, similar to the VTM task, we create a decoder query with a \texttt{[CLS]} embedding ($Q=1$), then 
apply a classification head with cross-entropy loss.

\section{Efficiency Analysis}
\label{sup_sec:efficiency}

We continue \cref{sec:efficiency} in the main paper to present an efficiency analysis on \vqa{} task.

\subsection{Scaling Latent Array}
\label{sup_sec:latent_scale}
\methodname{} has a complexity of $O(MN)$, while the input size $M$ is fixed for specific tasks and datasets.
In Fig.~\ref{fig:plot_fl_vqa}, we show the effect of varying the size of the latent array $N$ during finetuning in terms of computation and downstream \vqa{} performance.
Note that we use $N{=}128$ during pretraining.
We can see that the computational cost (GFLOPs) linearly scales with $N$, while the \vqa{} accuracy remains reasonably well
(e.g., CC+Webvid PT: $66.6 \rightarrow 68.0 \rightarrow 69.2 \rightarrow 69.8$ with latent array length $32 \rightarrow 64 \rightarrow 128 \rightarrow 256$), across three different pretraining setups (Sec.~\ref{supp_sec:ablation_pretraining_datasets}).

\begin{figure}[t]
  \centering
  \includegraphics[width=.45\textwidth]{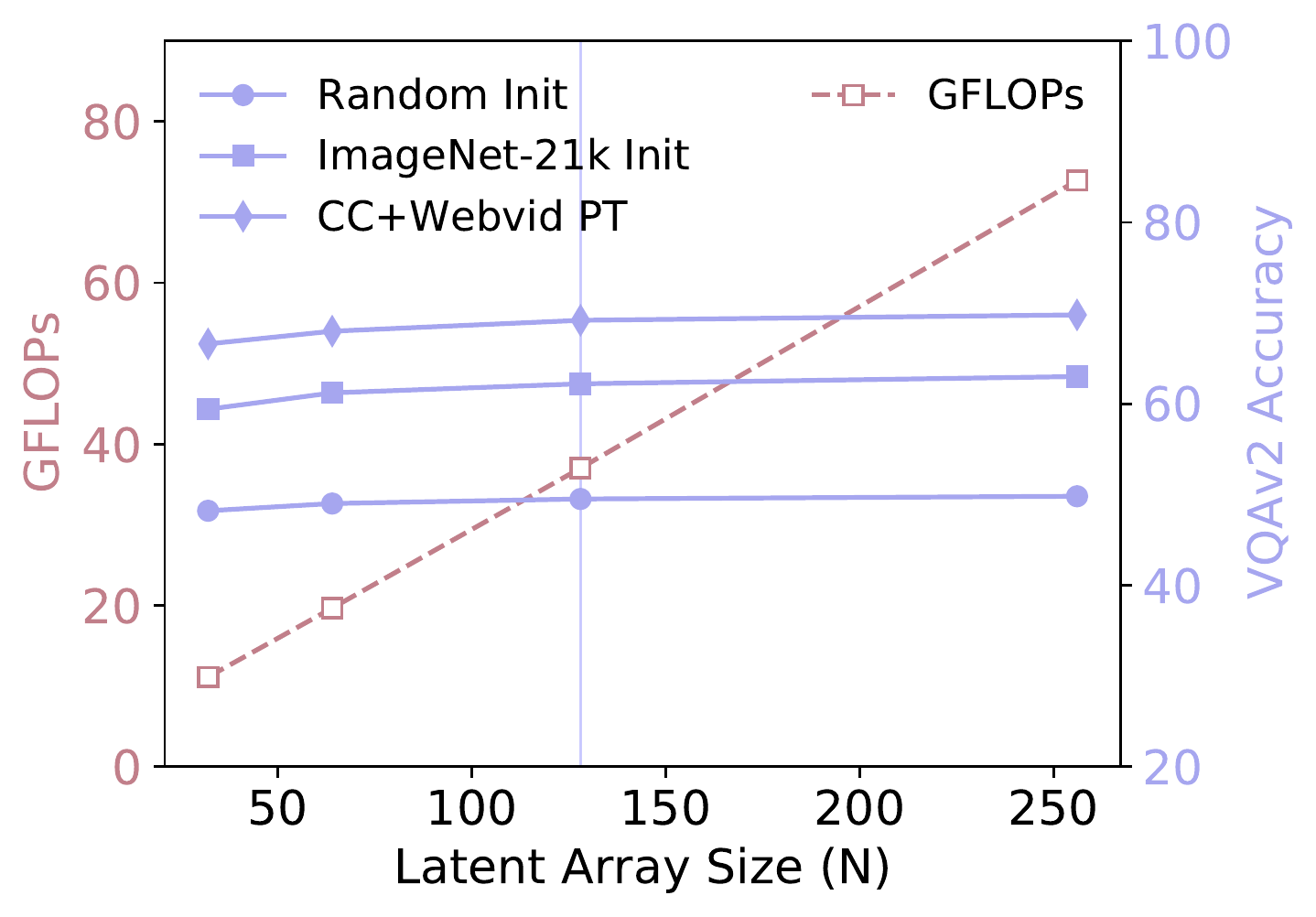}
  \caption{
  The efficiency-accuracy tradeoff of using different latent array size $N$ during finetuning on \vqa{}.
  During pretraining, we use the latent array size $N=128$ (blue vertical line).
  }
  \label{fig:plot_fl_vqa}
\end{figure}

\begin{table}[h]
\setlength{\tabcolsep}{6pt}
  \begin{center}
\resizebox{0.45\textwidth}{!}{
    {\small{
\begin{tabular}{ccc cc}
\toprule
\multicolumn{3}{c}{\# Cross-attentions in encoder} & \multirow{2}{*}{\vqa{} Acc.} & Time\\
\cmidrule{1-3}
Pretraining & Finetuning & Inference  & & (ms) \\
\midrule
3   & 3 & 3 & 68.7 & 18.0\\
$1 \sim 3$ ($0.5$) & 3 & 3 & 69.2 & 18.0\\
$1 \sim 3$ ($0.7$) & 3 & 3 & 68.9 & 18.0\\
\midrule
$1 \sim 3$ ($0.5$) & 1 & 1 & 68.2 & 15.0\\
$1 \sim 3$ ($0.5$) & 3 & 3 & 69.2 & 18.0\\
$1 \sim 3$ ($0.5$) & 3 & 1 & 66.1 & 15.0\\
\midrule
$1 \sim 3$ ($0.5$) & $1 \sim 3$ ($0.5$) & 1 & 68.4 & 15.0\\
$1 \sim 3$ ($0.5$) & $1 \sim 3$ ($0.5$) & 3 & 69.5 & 18.0\\
\bottomrule
\end{tabular}
}}}
\end{center}
\caption{
Accuracy and inference time on \vqa{} with varied number of cross-attentions in \methodname{} encoder.
We include the layer dropout probability $p^{LD}$ in brackets if used.
Note that \methodname{} has 3 cross-attention layers in encoder, and we do not apply dropout to the first cross-attention in encoder ($p^{LD}=0$) to ensure that the latent array always receives signal from the input.
}
\label{tab:layerdrop_vqa}
\end{table}

\subsection{LayerDrop to Encoder Cross-Attentions}
\label{sup_sec:layerdrop}

In Table~\ref{tab:layerdrop_vqa} we analyze the effect of applying LayerDrop (LD)~\cite{fan2019reducing} to encoder cross-attention layers, as discussed in main paper Sec.~\ref{sec:layerdrop} on an additional task, \vqa{}.
First, we observe that LD acts as a regularizer, as we see LD improves the \vqa{} accuracy in the first block, while increasing $p^{LD}$ too high $0.5\rightarrow0.7$ hurts the performance (69.2 $\rightarrow$ 68.9).
The last row in the bottom block achieves the best accuracy (69.5), with LD during both pretraining and finetuning.
Second, removing cross-attention layers without LD during finetuning hurts performance (see 69.2 $\rightarrow$ 66.1 in the middle block).
Lastly, with LD during finetuning, one can reduce the inference time latency around 16.7\% (18.0 ms $\rightarrow$ 15.0 ms), with minimal accuracy drop (see 69.5 $\rightarrow$ 68.4 in the bottom block).
This indicates that, with an LD-finetuned model, we can control its latency on demand at the inference time by varying the number of cross-attention layers, without storing checkpoints of multiple models.

\section{Ablation Studies}
\label{supp_sec:ablation}

We provide ablation studies regarding \methodname{}'s architectural components and training strategy, including
modality aggregation,
pretraining dataset,
positional encoding for latent arrays,
and 
two-stage training for CLIP weight initialization.

\subsection{Modality Aggregation}
\label{supp_sec:ablation_modality_aggregation}

In Table \ref{tab:ablation_aggre}, we compare different modality aggregation schemes for fusing visual and text inputs as we discussed in the main paper Sec.~\ref{sec:encoding}. This study is performed on \vqa{} with two different weight initializations.
In our experiments, we do not observe a significant difference among the three methods (\textit{Joint}, \textit{Separate}, \textit{Separate+}) in terms of accuracy and GFLOPs.
Thus, we use \textit{Joint} as our default modality aggregation scheme for simplicity.

\begin{table}[t]
\setlength{\tabcolsep}{6pt}
  \begin{center}
\resizebox{0.48\textwidth}{!}{
\begin{tabular}{lccc}
\toprule
\multirow{2}{*}{Aggregation Scheme} & \multicolumn{2}{c}{Weight initialization} & \multirow{2}{*}{GFLOPs $\downarrow$} \\
\cmidrule{2-3}
& Random init & \imagenet{} (ViT-B/32) & \\
\midrule
\textit{Joint} (default) & 48.6 & 62.5 & \textbf{30.5} \\
\textit{Separate} & 49.5 & 62.3 & 31.3\\
\textit{Separate+} & \textbf{50.5} & \textbf{62.9} & 33.2\\
\bottomrule
\end{tabular}}
\end{center}
\caption{
Comparison of different modality aggregation schemes (main paper Sec.~\ref{sec:encoding}) on \vqa{}.
}
\label{tab:ablation_aggre}
\end{table}

\subsection{Pretraining Datasets}
\label{supp_sec:ablation_pretraining_datasets}

Table \ref{tab:ablation_datasets} shows the ablation of pretraining datasets in terms of two downstream tasks, \vqa{} and \msrvtt{}.
Initializing \methodname{} parameters with ViT-B/32 \imagenet{} pretrained weights (main paper Sec.~\ref{sec:pretraining_setup}) greatly improves the performance over random initialization.
Further pretraining on image-text (\cc{}) or video-text (\webvid{}) datasets further improves the performance.
One interesting observation is that, pretraining on the data of the same format as the downstream task has slightly more advantages over data of different format -- compared to video-text data, pretraining on image-text data gives more performance gain on image-text task (\vqa{}), and vice versa.
The best performance is achieved by \methodname{} pretrained on both datasets, showing that our framework benefits from input data from both formats.

\begin{table}[t]
\setlength{\tabcolsep}{3pt}
\begin{center}
\resizebox{\linewidth}{!}{
\begin{tabular}{lccccc}
\toprule
\multirow{2}{*}{Pretraining Datasets} & \multicolumn{3}{c}{Modality} & \vqa{} & \msrvtt{} \\
\cmidrule(r){2-4} \cmidrule(r){5-5} \cmidrule{6-6}
& Image & Video & Text & Acc & R@1 \\
\midrule
Random Init (Standard Gaussian) &  & & & 48.6 & 6.2 \\
\imagenet{} (ViT-B/32) & \checkmark & & & 62.3 & 12.1\\
\imagenet{} (ViT-B/32) + \cc{} & \checkmark & & \checkmark & 68.2  & 24.6 \\
\imagenet{} (ViT-B/32) + \webvid{} & & \checkmark & \checkmark & 67.5 & 25.1 \\
\imagenet{} (ViT-B/32) + \cc{} + \webvid & \checkmark & \checkmark & \checkmark & \textbf{69.2} & \textbf{26.8} \\
\bottomrule
\end{tabular} 
}
\end{center}
\caption{
Comparison of different pretraining datasets on \vqa{} and \msrvtt{}.
\imagenet{} (ViT-B/32) refers to weight initialization from the ViT-B/32 checkpoint pretrained on \imagenet{} (main paper Sec.~\ref{sec:pretraining_setup}).
}
\label{tab:ablation_datasets}
\end{table}

\subsection{Learned vs. Fourier Positional Encodings for Latent Array}
\label{supp_sec:ablation_latent_position_encoding}

In Table~\ref{tab:ablation_pos}, we compare the learned~\cite{gehring2017convolutional,vaswani2017attention} and Fourier feature~\cite{stanley2007compositional,mildenhall2020nerf,jaegle2021perceiver1} positional encodings on \vqa{}, as discussed in main paper Sec.~\ref{sec:encoding}.
We do not see a meaningful difference between the two positional encodings in two different weight initialization settings.
Thus, we simply use the learned positional encoding as the default positional encoding for the latent array.

\begin{table}[t]
\setlength{\tabcolsep}{6pt}
\begin{center}
\resizebox{0.45\textwidth}{!}{
\begin{tabular}{lcc}
\toprule
\multirow{2}{*}{Positional Encoding} & \multicolumn{2}{c}{Weight init}\\
\cmidrule{2-3}
& Random Init & \imagenet{} (ViT-B/32) \\
\midrule
Learned (default) & 49.5 & \textbf{62.3} \\
Fourier & \textbf{49.7} & 62.2 \\
\bottomrule
\end{tabular}}
\end{center}
\caption{
Comparison of different position encodings for latent array on \vqa{}.
}
\label{tab:ablation_pos}
\end{table}

\begin{table}[h]
\begin{center}
\resizebox{0.6\linewidth}{!}{
\begin{tabular}{lc}
\toprule
Weight init & \msrvtt{} R@1\\
\midrule
One-stage & 36.3 \\
Two-stage (default) & \textbf{45.9} \\
\bottomrule
\end{tabular}}
\end{center}
\caption{
Comparison of one-stage vs. two-stage training for CLIP weight initialization on \msrvtt{}.
}
\label{tab:two_stage_train}
\end{table}

\begin{table*}[h]
\begin{center}
\setlength{\tabcolsep}{4pt}
\resizebox{\textwidth}{!}{
\begin{tabular}{lllcccccccc}
\toprule
\multirow{2}{*}{Model} & 
\multirow{2}{*}{Pretraining Datasets} & \multirow{2}{*}{Visual Backbone} & 
\multicolumn{4}{c}{Text-to-Video Retrieval (R@1/R@5/R@10) $\uparrow$} &
\multicolumn{2}{c}{QA Accuracy $\uparrow$} &
\multirow{2}{*}{GFLOPs $\downarrow$} &
\multirow{2}{*}{Time (ms) $\downarrow$}
\\
\cmidrule(lr){4-7}  \cmidrule(lr){8-9}
& & & \msrvtt{} & \didemo{} & \lsmdc{} & \activitynet{} & \tgifqa{} (A/T/F) & \msrvttqa{} & & \\

\midrule
\multicolumn{3}{l}{\textcolor{gray}{Models using other input modalities (\eg, audio)}} \\
\textcolor{gray}{\hero{} \cite{li2020hero}} & \textcolor{gray}{TV/\htm{}} 
& \textcolor{gray}{ResNet152+Slowfast \cite{he2016deep,feichtenhofer2019slowfast}}& \textcolor{gray}{20.5} / \textcolor{gray}{47.6} / \textcolor{gray}{60.9} & \textcolor{gray}{-} & \textcolor{gray}{-} & \textcolor{gray}{-} & \textcolor{gray}{-} & \textcolor{gray}{-} & \textcolor{gray}{935.2} & \textcolor{gray}{2200.0} \\
\textcolor{gray}{\mmt{} \cite{gabeur2020multi}} & \textcolor{gray}{\htm{} }
& \textcolor{gray}{S3D+VGG+DenseNet161 \cite{xie2018rethinking,hershey2017cnn,huang2017densely}} & \textcolor{gray}{26.6} / \textcolor{gray}{57.1} / \textcolor{gray}{67.1} & \textcolor{gray}{-} & \textcolor{gray}{12.9} / \textcolor{gray}{29.9} / \textcolor{gray}{40.1} & \textcolor{gray}{-} & \textcolor{gray}{-} &  \textcolor{gray}{-} & \textcolor{gray}{-} & \textcolor{gray}{-}\\
\textcolor{gray}{\alvnet{} \cite{rouditchenko2020avlnet}} & \textcolor{gray}{\htm{} }
& \textcolor{gray}{ResNet152+ResNeXt \cite{he2016deep,xie2017aggregated}} & \textcolor{gray}{27.1} / \textcolor{gray}{55.6} / \textcolor{gray}{66.6} & \textcolor{gray}{-} & \textcolor{gray}{17.0} / \textcolor{gray}{38.0} / \textcolor{gray}{48.6} & \textcolor{gray}{-} & \textcolor{gray}{-} & \textcolor{gray}{-} & \textcolor{gray}{153.4} & \textcolor{gray}{2000.0}\\
\midrule

\multicolumn{3}{l}{\textcolor{gray}{Models with CLIP initialization}} \\
\textcolor{gray}{Hunyuan \cite{min2022hunyuan_tvr}} & -
& \textcolor{gray}{CLIP (ViT-B/16)} &
\textcolor{gray}{\textbf{55.0}} / \textcolor{gray}{\textbf{80.4}} / \textcolor{gray}{\textbf{86.8}} & \textcolor{gray}{\textbf{52.1}} / \textcolor{gray}{\textbf{78.2}} / \textcolor{gray}{\textbf{85.7}} & \textcolor{gray}{\textbf{29.7}} / \textcolor{gray}{46.4} / \textcolor{gray}{55.4} &
\textcolor{gray}{\textbf{57.3}} / \textcolor{gray}{\textbf{84.8}} / \textcolor{gray}{93.1} & \textcolor{gray}{-} & \textcolor{gray}{-} & \textcolor{gray}{2022.8} & \textcolor{gray}{-}
\\
\textcolor{gray}{CLIP2TV \cite{gao2021clip2tv}} & \textcolor{gray}{-}
& \textcolor{gray}{CLIP (ViT-B/16)} & \textcolor{gray}{49.3} /  \textcolor{gray}{74.7} / \textcolor{gray}{83.6} & \textcolor{gray}{45.5} /  \textcolor{gray}{69.7} / \textcolor{gray}{80.6} & - & \textcolor{gray}{44.1} /  \textcolor{gray}{75.2} / \textcolor{gray}{\textbf{98.4}} & - & \textcolor{gray}{-} & \textcolor{gray}{2212.3} & \textcolor{gray}{-}
\\

\textcolor{gray}{DRL \cite{wang2022disentangled}} & \textcolor{gray}{-}
& \textcolor{gray}{CLIP (ViT-B/32)} & \textcolor{gray}{47.4} / \textcolor{gray}{74.6} / \textcolor{gray}{83.8} & \textcolor{gray}{49.0} / \textcolor{gray}{76.5} / \textcolor{gray}{84.5} & \textcolor{gray}{26.5} / \textcolor{gray}{\textbf{47.6}} / \textcolor{gray}{\textbf{56.8}} & \textcolor{gray}{46.2} / \textcolor{gray}{77.3} / \textcolor{gray}{88.2} & \textcolor{gray}{-} & \textcolor{gray}{-} & \textcolor{gray}{511.0} & \textcolor{gray}{320.0}
\\
\textcolor{gray}{CAMoE(+DSL) \cite{cheng2021improving}} & \textcolor{gray}{-} 
& \textcolor{gray}{CLIP (ViT-B/32)} & \textcolor{gray}{47.3} / \textcolor{gray}{74.2} / \textcolor{gray}{84.5}  & \textcolor{gray}{-} & \textcolor{gray}{25.9} / \textcolor{gray}{46.1} / \textcolor{gray}{53.7} & \textcolor{gray}{-} & \textcolor{gray}{-} & \textcolor{gray}{-} & \textcolor{gray}{399.7} & \textcolor{gray}{-}
\\
\textcolor{gray}{MDMMT-2 \cite{kunitsyn2022mdmmt}} & \textcolor{gray}{-}
& \textcolor{gray}{CLIP (ViT-B/32)} & \textcolor{gray}{48.5} / \textcolor{gray}{75.4} / \textcolor{gray}{83.9} & \textcolor{gray}{-} & \textcolor{gray}{26.9} / \textcolor{gray}{46.7} / \textcolor{gray}{55.9} & \textcolor{gray}{-} & \textcolor{gray}{-} & \textcolor{gray}{-} & \textcolor{gray}{-} & \textcolor{gray}{-}
\\
\textcolor{gray}{Ours$^{N=32, \text{Multi}}$ } & \textcolor{gray}{\cc{}+\webvid{}}
& \textcolor{gray}{CLIP (ViT-B/16)} & \textcolor{gray}{45.9} / \textcolor{gray}{71.0} / \textcolor{gray}{82.1} & \textcolor{gray}{-} & \textcolor{gray}{-} & \textcolor{gray}{-} & \textcolor{gray}{-} & \textcolor{gray}{-} & \textcolor{gray}{\textbf{80.0}} & \textcolor{gray}{\textbf{80.0}} \\

\midrule
\htm{} \cite{miech2019howto100m} & \htm{} 
& ResNet152+ResNeXt \cite{he2016deep,xie2017aggregated} & 14.9 / 40.2 / 52.8  & - & 7.1 / 19.6 / 27.9 & - & - & - & 164.3 & 1100.0\\
\clipbert{} \cite{lei2021less} & \coco{} / \cc{}
& ResNet50 \cite{Jiang2020} &  22.0 / 46.8 / 69.9 &  20.4 / 48.0 / 60.8 & - & 21.3 / 49.0 / 63.5 & 82.8 / 87.8 / 60.3 & 37.4 & 340.0 & 700.0 \\
\frozen{} \cite{bain2021frozen} & \cc{} / \webvid{}
& Timesformer-B/16 \cite{bertasius2021space} & 31.0 / 59.8 / 72.4 & \textbf{31.0} / \textbf{59.8} / 72.4 &  15.0 / 30.8 / 39.8 & - & - & - & 89.0 & 260.0 \\
Ours$^{N=128, \text{Mixed}}$  & \cc{} / \webvid{}
& ViT-B/32 \cite{dosovitskiy2020image} & \textbf{32.6} / \textbf{62.1} / \textbf{71.6} & 30.5 / 59.7 / \textbf{73.0} & \textbf{15.8} / \textbf{37.6} / \textbf{40.1} & \textbf{33.9} / \textbf{62.1} / \textbf{76.4} & \textbf{91.4} / \textbf{94.9} / \textbf{69.2} & \textbf{43.2} & \textbf{43.9} & \textbf{72.0}\\

\bottomrule
\end{tabular}
}
\end{center}
\caption{
Full metrics of finetuning performance on text-to-video retrieval and video question answering benchmarks.
We report R@1/R@5/R@10 for text-to-video retrieval tasks and report QA accuracy on the FrameQA task.
\textit{GFLOPs} shows the inference cost on a single sample, and  \textit{Time (ms)} indicates the average inference time across all samples on \msrvtt{} val split.
For a fair comparison, we gray out 1) the models that use input modalities other than video and text (\eg, audio) and 2) the models that use CLIP visual encoder \cite{radford2021learning} (the cross-attention layers of \methodname{} cannot be initialized with CLIP parameters and trained from scratch; see the discussion in Sec.~\ref{sec:overall_comparison}). $^{N=128}$ means latent size N=128.
$^{\text{Multi}}$ and $^{\text{Mixed}}$ mean multi-stream and mixed-stream respectively.
}
\label{tab:supp_downstream_videotext}
\end{table*}

\begin{table*}[h]
\setlength{\tabcolsep}{4pt}
\begin{center}
\resizebox{\textwidth}{!}{
\begin{tabular}{lllcccccc}
\toprule
\multirow{2}{*}{Model} & 
\multirow{2}{*}{Pretraining Datasets} & \multirow{2}{*}{Visual Backbone} & 
\multicolumn{1}{c}{Text-to-Image-to Retrieval $\uparrow$} &
\multicolumn{2}{c}{QA Accuracy $\uparrow$} &
\multirow{2}{*}{GFLOPs $\downarrow$} & 
\multirow{2}{*}{Time (ms) $\downarrow$}
\\
\cmidrule(lr){4-4}  \cmidrule(lr){5-6} 
& & & \flickrthirty{} (R@1/R@5/R@10) & \vqa{} & \nlvr{} (dev/test-P) & \\

\midrule

\multicolumn{3}{l}{\textcolor{gray}{Models using additional object tag inputs}} \\
\textcolor{gray}{\vinvl{} \cite{zhang2021vinvl}} & \textcolor{gray}{\coco{} / \cc{} / \sbu{} / \flickr{} / \oi{}*} & \textcolor{gray}{Faster-RCNN \cite{zhang2021vinvl}} & \textcolor{gray}{-} & \textcolor{gray}{75.95} & \textcolor{gray}{82.05} / \textcolor{gray}{83.08}  & \textcolor{gray}{1023.3} & \textcolor{gray}{800.0} \\
\textcolor{gray}{\oscar{} \cite{li2020oscar}} & \textcolor{gray}{\coco{} / \cc{} / \sbu{} / \flickr{}*}  & \textcolor{gray}{Faster-RCNN \cite{anderson2018bottom}} & \textcolor{gray}{-} & \textcolor{gray}{73.16} & \textcolor{gray}{78.07} / \textcolor{gray}{78.36} & \textcolor{gray}{956.4} & \textcolor{gray}{1000.0} \\
\midrule

\uniter{} \cite{chen2019uniter} & \coco{} / \cc{} / \sbu{} / \vg{} & Faster-RCNN \cite{anderson2018bottom} & \textbf{72.5} / \textbf{92.4} / \textbf{96.1} & \textbf{72.70} & 75.85 / 75.80 & 949.9 & 1000.0 \\
\vilt{} \cite{kim2021vilt} & \coco{} / \cc{} / \sbu{} / \vg{} & ViT-B/32 \cite{dosovitskiy2020image} &  64.4 / 88.7 / 93.8 & 71.26 & 75.70 / \textbf{76.13} & 55.9 & 32.0 \\
Ours$^{N=128}$ & \coco{} / \cc{} / \sbu{} / \vg{} & ViT-B/32 \cite{dosovitskiy2020image} & 62.4 / 87.1 / 93.2 & 71.62 & 75.45 / 75.53 & \textbf{30.5} & \textbf{18.0} \\

\midrule

\lxmert{} \cite{tan2019lxmert} & \coco{} / \vg{}* & Faster-RCNN \cite{anderson2018bottom} &  - & \textbf{72.42} & 94.90 / 74.50  & 952.0 & 1100.0 \\
\visualbert{} \cite{li2019visualbert} & \coco{} & Faster-RCNN \cite{anderson2018bottom} & - & 70.80 & 67.40 / 67.00  & 425.0 & 1000.0 \\
\pixelbert{} \cite{huang2020pixel} & \coco{} / \vg{} & ResNet50 \cite{he2016deep} &  53.4 / 80.4 / 88.5 & 71.35 & 71.70 / 72.40 & 136.8 & 150.0 \\

Ours$^{N=128}$ & \coco{} / \vg{} & ViT-B/32 \cite{dosovitskiy2020image} & \textbf{61.7} / \textbf{86.7} / \textbf{92.1} & 70.45 & \textbf{73.30} / \textbf{74.87} & \textbf{30.5} & \textbf{18.0} \\

\midrule
\frozen{} \cite{bain2021frozen} & \cc{} / \webvid{} &  Timesformer-B/16 \cite{bertasius2021space}  & 61.0 / 87.5 / 92.7 & - & -  & 63.9 & 70.0 \\

Ours$^{N=64}$ & \cc{} / \webvid{} & ViT-B/32 \cite{dosovitskiy2020image} & 61.0 / 86.6 / 93.0 & 70.12 & 74.04 / 74.52 & \textbf{17.0} & \textbf{8.0} \\
Ours$^{N=128}$ & \cc{} / \webvid{} & ViT-B/32 \cite{dosovitskiy2020image} & \textbf{61.8} / \textbf{88.0} / \textbf{92.9} & \textbf{70.91} & \textbf{75.30} / \textbf{75.44} & 30.5 & 18.0 \\
\bottomrule
\end{tabular} 
}
\end{center}
\caption{
Finetuning performance on text-to-image retrieval and visual question answering benchmarks.
For \nlvr{}, we show Test-P accuracy. For \flickrthirty{}, we show text-to-image retrieval R@1. Note that for brevity, we only show the image or video source datasets for \textit{Pretraining Datasets}; the datasets that added additional text annotations are not included in the column (we use * to highlight them). For example, \lxmert{} is trained with image-text datasets COCO and VG, as well as the three QA datasets based on COCO and VG images, \ie, \vqa{}, \vgqa{} and \gqa{}.
We also gray out models that use additional object tags in the first block and are not comparable to our model.
\textit{GFLOPs} shows the inference cost on a single sample, \textit{Time (ms)} indicates the average inference time over all samples in \vqa{} minival split. 
For a fair comparison, we gray out models that are pretrained with more data. $^{N=128}$ means latent size N=128. 
}
\label{tab:supp_downstream_imagetext}
\end{table*}

\subsection{Two-stage Training for CLIP Weight Initialization}
\label{supp_sec:two_stage_pretraing_clip_init}

In Table~\ref{tab:two_stage_train}, we compare the
two-stage and one-stage training for weight initialization form CLIP, as discussed in main paper Sec.~\ref{sec:weightinit}.
We use the architecture with latent size $N=32$.
We see significant improvement with two-stage training on \msrvtt{} and suggest the training strategy for weight initialization from transformer architecture such as CLIP.

\section{Full Experiment Results}
\label{sup_sec:full_experiment_results}

In \Cref{tab:supp_downstream_videotext} and \Cref{tab:supp_downstream_imagetext}, we provide the full experiment results with R@1/R@5/R@10 scores for retrieval tasks.
}

\end{document}